\pdfoutput=1
\documentclass[11pt]{article}
\usepackage{booktabs}
\usepackage{listings}
\usepackage{xcolor}

\usepackage{amsmath, amssymb}
\usepackage{hyperref}
\usepackage{enumitem}
\usepackage{multirow}
\setlist[itemize]{noitemsep, nolistsep}
\definecolor{codegreen}{rgb}{0,0.6,0}
\definecolor{codegray}{rgb}{0.5,0.5,0.5}
\definecolor{codepurple}{rgb}{0.58,0,0.82}

\lstdefinestyle{mystyle}{
    commentstyle=\color{codegreen},
    keywordstyle=\color{magenta},
    stringstyle=\color{codepurple},
    basicstyle=\ttfamily\footnotesize,
    breakatwhitespace=false,         
    breaklines=true,                 
    captionpos=b,                    
    keepspaces=true,                 
    showspaces=false,                
    showstringspaces=false,
    showtabs=false,                  
    tabsize=2
}
\usepackage[final]{ACL2023}

\usepackage{times}
\usepackage{latexsym}
\usepackage[T1]{fontenc}
\usepackage[utf8]{inputenc}
\usepackage{microtype}
\usepackage{inconsolata}

\title{PythonSaga: Redefining the Benchmark to Evaluate Code Generating LLMs}

\author{Ankit Yadav, Himanshu Beniwal, Mayank Singh \\
Department of Computer Science and Engineering\\
Lingo Research Group \\
IIT Gandhinagar, Gujarat, India \\
\texttt{\{ankityadav,himanshubeniwal,singh.mayank\}@iitgn.ac.in}}
\usepackage{graphicx}

\begin{document}

\maketitle

\begin{abstract}

Driven by the surge in code generation using large language models (LLMs), numerous benchmarks have emerged to evaluate these LLMs capabilities. We conducted a large-scale human evaluation of \textit{HumanEval} and \textit{MBPP}, two popular benchmarks for Python code generation, analyzing their diversity and difficulty. Our findings unveil a critical bias towards a limited set of programming concepts, neglecting most of the other concepts entirely. Furthermore, we uncover a worrying prevalence of easy tasks that can inflate model performance estimations. To address these limitations, we propose a novel benchmark, \texttt{PythonSaga}, featuring 185 hand-crafted prompts in a balanced representation of 38 programming concepts across diverse difficulty levels. The robustness of our benchmark is demonstrated by the poor performance of existing Code-LLMs. 
The code and data set are openly available to the NLP community at \url{https://anonymous.4open.science/r/PythonSaga}.

\end{abstract}

\section{Introduction}
\label{sec:intro}
\begin{figure}[!tbh]
    \centering
\includegraphics[scale=1.5]{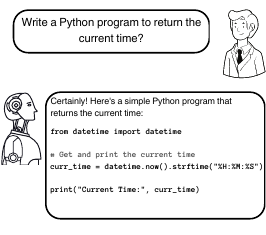}
    \caption{Illustration of a conversation wherein a human provides an input description, and a Code-LLM generates the expected Python code.}
    \label{fig:example}
\end{figure}

The rapid advancement of large language models (LLM), such as Gemini~\citep{team2023gemini}, GPT-4~\citep{openai2024gpt4}, LLaMA~\citep{touvron2023llama} and PaLM~\citep{anil2023palm}, has achieved near-human or even superhuman performance~\citep{bowman2023eight} in a wide range of NLP tasks. This surge has also led to the development of custom code generation models, such as Codex~\citep{chen2021evaluating}, STARCODER~\citep{li2023starcoder}, CodeGen~\citep{nijkamp2022codegen}, Code Llama~\citep{roziere2023code}, CodeGeeX~\citep{zheng2023codegeex}, and Deepseek Coder~\citep{guo2024deepseek}. These specialized models, collectively referred to hereafter as \textit{ ``Code-LLMs''}, harness the capabilities of LLMs for automated code generation from human descriptions. Figure~\ref{fig:example} shows a toy example with an input description from a human and an expected Python code generated by a Code-LLM. 

The prevalence of Python as the dominant programming language has significantly influenced the majority of Code-LLMs to showcase their code-generation capabilities on Python-specific benchmarks. Consequently, HumanEval~\citep{chen2021evaluating}, MBPP~\citep{austin2021program}, APPS~\citep{hendrycks2021measuring}, and DS-1000~\citep{lai2023ds} have emerged as prominent benchmarks, leveraging data curated from popular coding platforms like GitHub~\citep{GitHub}, LeetCode~\citep{GFG}, and Codeforces~\citep{Codeforces} and crowdsourcing efforts. These benchmarks offer a diverse range of programming challenges, with sizes ranging from a few hundred instances in HumanEval~\citep{chen2021evaluating}) to several thousand instances in datasets like APPS~\citep{hendrycks2021measuring} and MBPP~\citep{austin2021program}.

Code generation benchmarks, like their NLP counterparts~\citep{kiela2021dynabench}, are reaching saturation, revealing limitations in their ability to evaluate models comprehensively. Figure~\ref{fig:temporal_performance} present in appendix reports \textit{pass@1} score\footnote{\textit{pass@k} measures if at least one of the $k$ code samples generated by the model passes every test case. Detailed formal definition is present in Appendix~\ref{appendix:performance}.} of recent Code-LLMs on two popular benchmarks, HumanEval~\citep{chen2021evaluating} and MBPP~\citep{austin2021program}. This performance, approaching human-level capabilities, raises two key questions: \textbf{(1)} Have Code-LLMs attained the generalization ability to solve any programming problem? \textbf{(2)} What programming concepts remain challenging for them, hindering their ability to solve specific problems? Surprisingly, despite their widespread use, existing benchmarks lack a comprehensive evaluation of their diversity in terms of programming concepts and difficulty level.

In this paper, we introduce a comprehensive hierarchical classification of programming concepts, categorizing them into basic, intermediate, and advance levels  (see Section~\ref{sec:category}). We then rigorously evaluate two benchmarks, HumanEval~\citep{chen2021evaluating} and MBPP~\citep{austin2021program}, on two dimensions: diversity of programming concepts and user-perceived difficulty.  Our findings reveal a significant bias towards a small subset (<53\%) of programming concepts, leaving the vast majority underrepresented. Additionally, more than 80\% of the problems are perceived as easy, raising concerns about the generalizability and effectiveness (see Section~\ref{sec:exp}). To address these limitations, in Section~\ref{sec:new_dataset}, we propose a novel code generation benchmark, \texttt{PythonSaga}, featuring a balanced representation of 38 programming concepts across three difficulty levels in the form of 185 manually crafted problems. Surprisingly, our experiments show poor \textit{pass@1} scores by the majority of the existing open ($<4.5\%$) and closed-source ($<13\%$) Code-LLMs on \texttt{PythonSaga}. Furthermore, a detailed analysis uncovers significant disparities in their capacity to handle different programming concepts and difficulty levels. 

\section{Related Work}
\label{sec:rel_work}
\noindent \textbf{NLP for Programming:} Over the years, various programming tasks, including clone detection~\citep{roy2009comparison}, defect detection~\citep{tabernik2020segmentation}, code completion~\citep{hindle2016naturalness}, automated code repair~\citep{arcuri2008novel}, code search~\citep{sachdev2018retrieval}, and code summarization~\citep{allamanis2016convolutional}, have been extensively investigated and discussed within the NLP community. This exploration has led to the development of several datasets such as GitHub Java Corpus~\citep{allamanis2013mining}, BigCloneBench~\citep{svajlenko2014towards}, POJ-104~\citep{mou2016convolutional}, PY150~\citep{raychev2016probabilistic}, Devign~\citep{zhou2019devign}, Bugs2Fix~\citep{tufano2019empirical}, CodeSearchNet~\citep{husain2019codesearchnet}, CT-max/min~\citep{feng2020codebert}, MBPP by~\citet{austin2021program}, CodeXGLUE by~\citet{lu2021codexglue}, CodeNet by~\citet{puri2021codenet}, HumanEval by~\citet{chen2021evaluating},  XLCoST by~\citet{zhu2022xlcost}, MultiPL-E by~\citet{cassano2022multipl}, and  HumanEval-X by~\citet{zheng2023codegeex}. These datasets and associated benchmarks span multiple programming languages, including Java, C, C++, PHP, Ruby, Go, and Python, among others. 

\noindent \textbf{Code Generation Models:} The remarkable surge in the popularity of LLMs has also been accompanied by significant advancements in Code-LLMs. These models exhibit the capability to generate code in designated programming languages, guided by instructions presented in the form of prompts, functions, or docstrings. Prominent examples of such Code-LLMs include but are not limited to, Codex~\citep{chen2021evaluating}, CodeGen~\citep{nijkamp2022codegen}, Code Llama~\citep{roziere2023code}, STARCODER~\citep{li2023starcoder},  CodeGeeX~\citep{zheng2023codegeex}, Deepseek Coder~\citep{guo2024deepseek}, and StarCoder2~\citep{lozhkov2024starcoder}. These Code-LLMs are largely multilingual, capable of handling multiple programming languages, and their parameter sizes range from 1 billion to 70 billion. Their training datasets encompass popular programming websites and code repositories such as GitHub, LeetCode, and GeeksForGeeks. All popular Code-LLMs primarily focus on Python programs due to its widespread usage in ML and AI applications.

\noindent \textbf{Python-based Evaluation Benchmarks:} 
Recent thrust in Python code generation models also led to the development of several benchmark datasets. The PY150 dataset~\citep{raychev2016probabilistic}, consisting of 150,000 Python source files from GitHub, serves as a valuable tool for LLM evaluation. The APPS dataset~\citet{hendrycks2021measuring} features 10,000 problems from platforms like Codewars, AtCoder, Kattis, and Codeforces. HumanEval~\citep{chen2021evaluating} comprises 164 handwritten problems. The MBPP dataset~\citep{austin2021program} contains 974 entry-level problems. Additionally, the MathQA-Python dataset~\citep{austin2021program}, with 23,914 problems, assesses code synthesis from complex textual descriptions.

\begin{table*}[!tbh]
\centering
\begin{tabular}{ccc}  
\toprule
\textbf{Basic}& \textbf{Intermediate}& \textbf{Advance}\\ \toprule  
Function& OOPS
& Trie
\\
Mathematics& Stack
&Tree
\\ 
File Handling & Sorting
& Heap
\\ 
 Basic Libraries& Hashing
& Graph 
\\
Error Handling& Searching&Matrix 
\\ 
Input and Output& Recursion& 
Max Flow
\\
In-Built Functions& Linked List&Disjoint Set
\\ 
Pattern Replication&  Bit Manipulation& Backtracking
\\  
Basic Data Structures& Queue \& Dequeue& Greedy Search
\\ 
Variable \& Data Types& Regular Expression&Advanced OOPs
\\ 
Control Flow \& Conditions& Circular \& Doubly Linked List&Context Managers
\\ 
 & Advanced String Manipulation&Divide and Conquer
\\
 & &Dynamic Programming
\\
 & &Closures and Decorators
\\
 & &Concurrency and Parallelism\\\bottomrule
\end{tabular}
\caption{
A hierarchy of 38 programming concepts categorized into basic, intermediate, and advance categories.
}
\label{tab:concepts}
\end{table*}

\noindent \textbf{Limitations in Existing Benchmarks:} Current datasets for evaluating Large Language Models (LLMs) often lack transparency and comprehensiveness in problem selection and categorization. This opacity hinders assessments of the generalizability and representativeness of the benchmarks, potentially leading to overestimation of LLM performance on code generation tasks. To address this issue, this paper proposes a comprehensive problem categorization by outlining recommended concepts for problem inclusion, aiming to establish a rigorous and transparent benchmarking framework.

\section{Programming Concepts and Difficulty Levels}
\label{sec:category}
\subsection{Programming Concepts}
\label{sec:concepts}
The concepts encompass language-specific constructs like variables, data types, control flow, and conditions to generic constructs like Algorithms, OOPs, etc. We, therefore, propose a hierarchy of programming concepts in which a complex concept might require knowledge of several basic concepts. For example, sorting algorithms like \texttt{Quicksort} or \texttt{Mergesort} require a thorough understanding of data structures such as arrays and linked lists, as well as proficiency in algorithmic analysis and time complexity\footnote{\url{https://shorturl.at/nrBTX}}. Each programming concept is an intrinsic feature of a problem. We next describe the proposed hierarchy: 
\begin{itemize}
    \item \textbf{Basic Concepts:}  At the basic level, concepts involve the application of elementary syntax principles, encompassing the utilization of variables, manipulation of diverse data types, basic input/output operations, comprehension of control flow and conditional statements, basic handling of data structures, functions, and knowledge of essential built-in libraries. Problems leveraging basic concepts primarily aim to evaluate the core competencies within a designated programming language.
     \item \textbf{Intermediate Concepts}: Intermediate-level concepts involve a comprehensive understanding of multiple foundational concepts and their adept integration. For example, extending basic data structures to implement \texttt{Stack}, \texttt{Hash}, \texttt{Queue}, etc. Problems comprising intermediate concepts evaluate a higher level of proficiency in programming. 
    \item \textbf{Advance Concepts:} Concepts include implementation knowledge of advanced data structures such as \texttt{Tree}, \texttt{Heap}, etc., algorithmic paradigms such as \texttt{Greedy}, \texttt{Divide and Conquer}, and \texttt{Dynamic Programming}, and  \texttt{Concurrent and Parallel Programming}. Problems comprising advanced concepts focus on evaluating sophisticated problem-solving and design capabilities. 
\end{itemize}

We curate a list of 38 programming concepts from three popular coding platforms~\citep{GFG, LeetCode, hackerearth}. We further assign each concept to one of the three hierarchy levels. Table~\ref{tab:concepts} presents the curated concepts and the proposed hierarchy. 

\subsection{Difficulty Levels}
\label{sec:levels}
An annotator, with their expertise and experience in programming, can perceive a programming problem as belonging to one of three difficulty levels: \textit{Easy}, \textit{Medium}, or \textit{Hard}~\citep{hendrycks2021measuring}. Thus, difficulty level is an extrinsic feature of a problem. This subjective assessment is based on a complex combination of factors, such as knowledge of programming concepts, problem-solving skills, experience with similar problems, and coding proficiency. It is important to note that perceived difficulty is subjective and can vary significantly between annotators. A problem considered easy by one annotator due to their prior experience and knowledge might be deemed challenging by another who lacks those same advantages. Furthermore, the perceived difficulty of a problem can also evolve over time as an annotator develops their skills and knowledge. A problem that initially seemed challenging may become easier with practice and exposure to similar problems. Conversely, an annotator may encounter a problem that initially appears straightforward but then finds themselves struggling due to hidden complexities or unforeseen challenges.

In this paper, we focus on Python Programming language and conduct human experiments with two popular Python-based code generation benchmarks to showcase extensive selection bias and poor diversity in the curated problems. The following section describes the human experiments in detail.

\section{Limitations of Existing Code-Generation Benchmarks}
\label{sec:exp}

\subsection{Python Code Generation Benchmarks}
\label{sec:py_bench}
This study is grounded on the two most widely recognized Python code generation benchmarks: (i) HumanEval~\citep{chen2021evaluating} and (ii) MBPP~\citep{austin2021program}. Recent Code-LLMs including STARCODER~\citep{li2023starcoder}, LLaMA~\citep{touvron2023llama}, METAGPT~\citep{hong2023metagpt}, Code Llama~\citep{roziere2023code},  SANTACODER~\citep{allal2023santacoder}, CodeGeeX~\citep{zheng2023codegeex}, Gemma~\citep{team2024gemma}, and Claude 3~\citep{anthropic2024claude} have employed these two benchmarks to assess their performance. We next briefly describe these benchmarks.

\begin{itemize}
    \item \textbf{HumanEval Dataset:} HumanEval dataset was introduced alongside Codex~\citep{chen2021evaluating}\footnote{Codex is a GPT-based language model fine-tuned on publicly available codes from GitHub.}. It comprises 164 hand-crafted Python programming problems\footnote{Dataset is available here: \url{https://github.com/openai/human-eval}}. Each problem description contains a function signature, docstring, body, and multiple unit tests. Figure~\ref{fig:humaneval_example} illustrates a representative problem. On average, each problem is associated with 7.7 unit tests.  
    
    \item \textbf{Mostly Basic Programming Problems (MBPP) Dataset:} The MBPP dataset~\citep{austin2021program} evaluates models that can synthesize short Python programs from natural language descriptions. The benchmark\footnote{Dataset is available here: \url{https://github.com/google-research/google-research/tree/master/mbpp}} consists of about 974 crowd-sourced Python programming problems designed to be solvable by entry-level programmers. Each problem consists of a task description, code solution, and three automated test cases. 
    Figure~\ref{fig:mbpp_example} presents a representative problem.

\end{itemize}
Both benchmarks evaluate model performances against one of the most popular metrics \textit{pass@k}. We formally define \textit{pass@k} in appendix~\ref{appendix:performance}. 

\subsection{Human Annotation Experiments}
\label{sec:human_anno}
Next, we conducted two human annotation studies to gain insights into the diversity in programming concepts and difficulty levels of the two proposed benchmarks. Each study involved the recruitment of the same set of five annotators. Each annotator is a postgraduate student in Computer Science with at least three years of experience in Python programming and competitive programming. It is noteworthy that each participant willingly volunteered throughout the entire duration of the experiment, and no remuneration was provided. Internet access was prohibited during the entire annotation period. Annotators were encouraged to utilize any brute-force technique they considered appropriate without prioritizing optimized solutions. No time constraints were imposed to prevent hasty or fatigue-induced decisions. Each annotator was presented with 164 problems from HumanEval and a randomly selected set of 164 problems from MBPP.
We next describe the two annotation studies:

\begin{itemize}
    \item \textbf{Programming Concepts Diversity:} In this study, we adopted a single-concept annotation approach, where annotators assigned one programming concept (detailed in Section~\ref{sec:concepts}) to each problem. This selection represented the concept they considered most crucial for successful problem-solving. Our annotation guidelines explicitly prohibited assigning multiple concepts to any single problem, ensuring focused and unambiguous mapping between problems and relevant concepts.
    
    \item \textbf{Difficulty Level Diversity:} In this study, each annotator categorized the problems into three distinct difficulty levels: \textit{Easy}, \textit{Medium}, and \textit{Hard}, based on their individual expertise and experience. 
\end{itemize}

\begin{figure}[!tbh]
\centering
\includegraphics[trim={2cm 4cm 0 2cm},clip, scale=0.5]{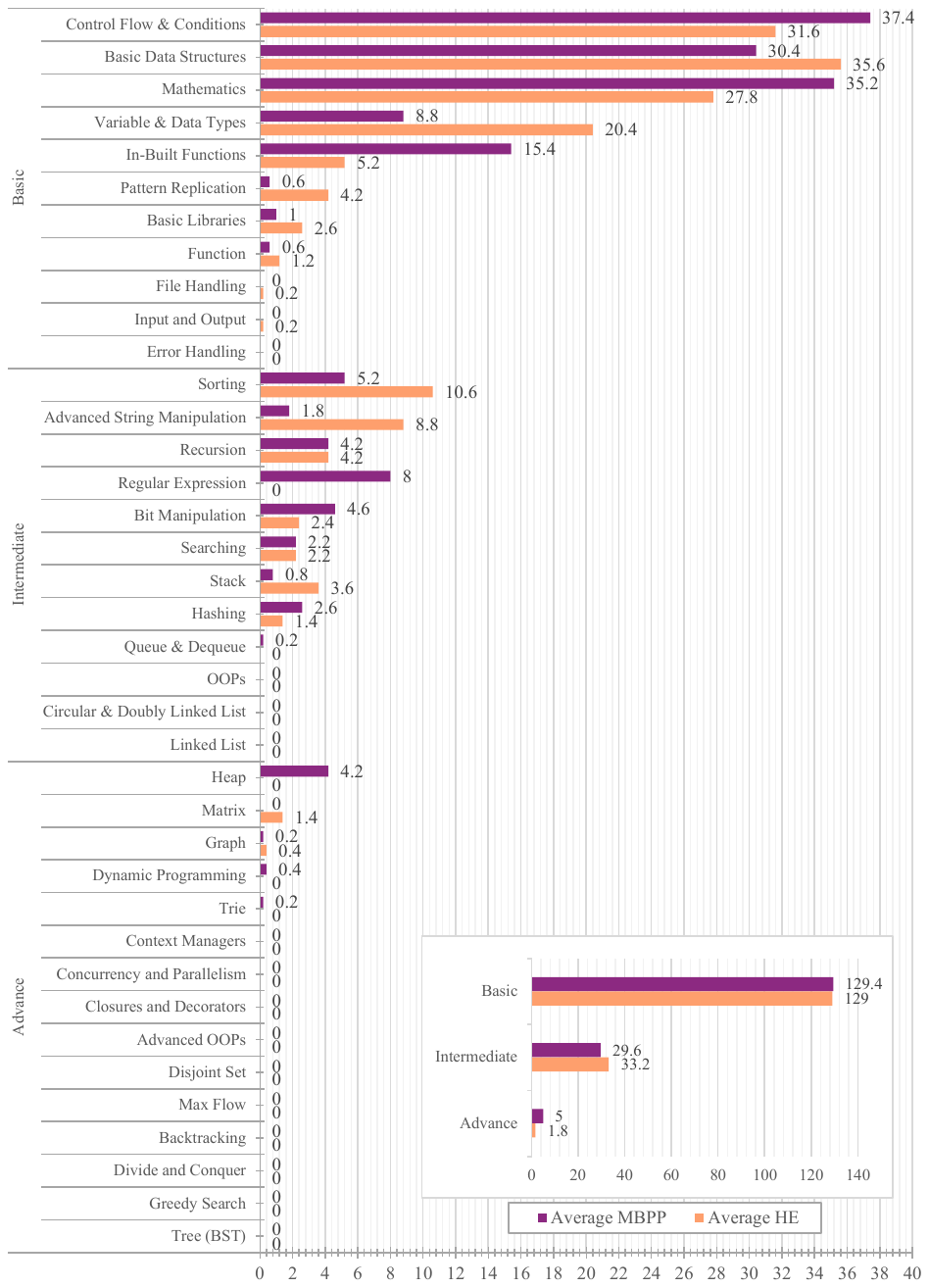}
\caption{Average number of problems in each of the programming concepts for two benchmarks, HumanEval and MBPP. The average number of problems assigned to each programming concept was determined by averaging the concept labels provided by five independent annotators. }\label{fig:diversity_concepts}
\end{figure}

\subsection{Observations}
\label{sec:human_observ}

\noindent \textbf{Diversity in the Programming Concepts:}
In this section, we report the proportion of problems assigned to a specific concept averaged over five annotators. We find five predominant concepts, \textit{Mathematics}, \textit{Control Flow and Conditions}, \textit{Basic Data Structures}, \textit{Variables and Data Types}, and \textit{In-Built Functions}, which comprise 72.1\% and 77.3\% problems in HumanEval and MBPP, respectively. Surprisingly, we found a complete absence of 14 (=37.8\%) concepts.  Notable exclusions include \textit{OOPs}, \textit{Linked-lists}, \textit{Tree}, \textit{Graph}, and \textit{Backtracking}. Figure~\ref{fig:diversity_concepts} presents conceptwise proportions in both the benchmarks. Further analysis suggests that, on average, the Basic category comprises approximately 78\% of problems in both HumanEval and MBPP. The Intermediate category comprises 20.24\%  and 18.04\% problems in HumanEval and MBPP, respectively. Finally, the Advance category contains  1.09\% and 3.04\% problems in HumanEval and MBPP, respectively.
          
\noindent \textbf{Diversity in the Difficulty level:}
Here, we report the difficulty level assigned to a problem using majority voting among the annotators. In HumanEval, 84.8\% of the problems were classified as \textit{Easy}, 14.6\% as \textit{Medium}, and only 0.6\% as \textit{Hard}. Whereas in MBPP,  89.6\% and 10.4\% of problems were categorized as \textit{Easy}, and \textit{Medium}, respectively. No problem in MBPP was labeled as \textit{Hard}. Here, we achieved significant consensus among the annotators. For example, in HumanEval, we find complete agreement among five annotators on 39\% of the problems. Whereas we miss complete agreement by a single vote in 29.2\% problems. In the case of MBPP, the 40.2\% problems resulted in a complete agreement, with 42.1\% problems missing the complete agreement by one vote. 

We notice a considerable selection bias favoring certain programming concepts and simpler problems in both benchmarks. Consequently, we assert that current benchmarks overstate the generalization abilities of existing code-LLMs.

\section{\texttt{PythonSaga}: A New Benchmark for Code Generation Models}
\label{sec:new_dataset}

We now introduce \texttt{PythonSaga}, a new Python code generation benchmark that addresses the limitations of existing benchmarks in terms of diversity in concepts and difficulty level. \texttt{PythonSaga} contains 185 prompts, close to equal representation from each of the 38 programming concepts with varied levels of difficulty (described in Section~\ref{sec:levels}).

\subsection{Data Sources and Curation Methodology}
\label{sec:saga_sources}
Aligned with the problem curation strategies employed in established benchmarks~\citet{hendrycks2021measuring,lai2023ds,zhu2022xlcost}, we curate problems from two prominent coding platforms: GeekForGeeks~\citep{GFG} and LeetCode~\citep{LeetCode}. To comprehensively represent each proposed programming concept (detailed in Section~\ref{sec:concepts}), we curated five problems per concept. This diverse set comprises one \textit{Easy} problem, two \textit{Medium} problems, and two \textit{Hard} problems, ensuring a balanced distribution across difficulty levels (20\%, 40\%, and 40\%, respectively) within the \texttt{PythonSaga} Dataset.The difficulty levels for each question were specified by their respective source platforms, GeekForGeeks, and LeetCode.

\begin{table}[!tbh]
\centering
\resizebox{\hsize}{!}{  
\begin{tabular}{lccl}  
\toprule
 \textbf{Model}& \textbf{Size}& \textbf{Pass@1} &\textbf{Pass@10}\\\toprule
  StarCoderBase&  7B& 0.0029&0.0149\\
 StarCoder2& 7B& 0.0024&0.0217\\
 Code Llama& 7B& 0.0067&0.0472\\
 CodeQwen1.5-Chat& 7B& 0.0059&0.0497\\
 Nxcode-CQ-orpo& 7B& 0.0058&0.0523\\
 Mistral-Instruct-v0.1 & 7B& 0.0140&0.0552\\
 Code Llama Instruct& 7B& 0.0178&0.0744\\
 Deepseek Coder Instruct & 6.7B& 0.0137&0.0889\\
 Code Llama Python& 7B&   0.0240&0.0979\\
 Llama 3& 8B& 0.0370&0.1125\\
 Phi-2& 2.7B& 0.0302&0.1187\\
 OpenCodeInterpreter-DS& 6.7B& 0.0259&0.1206\\ 
 Deepseek Coder& 6.7B&  0.0343&0.1415\\
 Code Llama Python& 13B& 0.0405&0.1514\\ 
  \color{gray}GPT-3.5& \color{gray} NA&   \color{gray}0.0724& \color{gray}0.2384\\
   \color{gray}GPT-4&  \color{gray}NA&   \color{gray}0.1243& \color{gray}0.3311\\ 

\bottomrule
\end{tabular}}
\caption{Comparison between open and closed-source models on \texttt{PythonSaga}. We use the number of samples ($n$) as 20 all models. }
\label{tab:open-close-pythonsaga}
\end{table}

To enhance human-friendliness and ground the problems in realistic contexts, each shortlisted problem statement undergoes a manual rephrasing process without any aid from AI tools. Furthermore, a comprehensive description of input and output formats, accompanied by relevant examples, is supplied with each problem statement to ensure a thorough understanding of the task by Code-LLM. This multi-step approach aims to retain the core knowledge and essential solution steps while integrating them into relatable real-world scenarios. This reconstruction involves reformulating the entire problem statement while preserving its fundamental functionality. This deliberate transformation enhances the challenge for Code-LLMs, requiring them to move beyond simple pattern matching and grasp the nuanced context embedded within the prompt to devise a solution effectively. For example, the problem statement \textit{``Given a string str, find the length of the longest substring without repeating characters.''} is paraphrased as \textit{``Let's say you attend a car show where cars of different brands are showcased in a row. Find the length of the longest stretch where no two cars are of the same brand. Take the input from the user for the brands of the cars in the order they are placed in the row. Print the length of the longest stretch where no two cars are of the same brand''}.

\subsection{Size and Structure}
\label{sec:saga_size_nomenclature}
Overall, \texttt{PythonSaga} comprises five problem instances from each programming concept, resulting in a total size of 185 problems. Each problem is associated with a maximum of four test cases, with an average of 3.7 test cases per problem. \texttt{PythonSaga}'s structure resembles  HumanEval and MBPP, wherein each problem comprises a function signature, docstring, body, and multiple unit tests. A representative example is present in Appendix~\ref{sec:appex_rep_PythonSaga}.

\subsection{Benchmarking Existing LLMs}
\label{sec:saga_per_exist_models}
Next, we benchmark several open and closed-source LLMs on \texttt{PythonSaga}. Open-source models include StarCoderBase~\citep{li2023starcoder}, StarCoder2~\citep{lozhkov2024starcoder}, Code Llama~\citep{roziere2023code}, CodeQwen1.5-Chat~\citep{qwen}, Nxcode-CQ-orpo~\citep{Nxcode-CQ}, Mistral-Instruct-v0.1~\citep{jiang2023mistral}, Code Llama Instruct~\citep{roziere2023code}, Deepseek Coder Instruct~\citep{guo2024deepseek}, Code Llama Python~\citep{roziere2023code} 7B \& 13B, Llama 3~\citep{meta2024introducing}, Phi-2~\citep{javaheripi2023phi}, OpenCodeInterpreter-DS~\citep{zheng2024opencodeinterpreter}, and Deepseek Coder~\citep{guo2024deepseek}. Except for Mistral-Instruct-v0.1, the rest are Code-LLMs. In addition, we benchmark two closed-source models, including GPT variants GPT-3.5~\citep{chatGPT3.5} and GPT-4~\citep{openai2024gpt4}. While larger open-source options exist, our selection was restricted to models with  less than 13B parameters due to computational resource limitations, which were limited to a single Tesla V100 in our case.

We evaluate model performances using \textit{pass@k} metric. Adhering to previous studies like HumanEval~\citep{chen2021evaluating}, StarCoder~\citep{li2023starcoder}, Deepseek Coder~\citep{guo2024deepseek} etc, we primarily utilized $k=1$, signifying that a model is considered successful if at least one of its generated solutions passes the defined evaluation criteria. However, we additionally explored $k=10$ to analyze model consistency across larger sets of responses. Notably, unlike prior works that varied the number of sampled responses ($n$), we consistently generated $n=20$ samples from both open and closed source models for a consistent evaluation.

\begin{figure*}[!tbh]
    \includegraphics[width=\textwidth]{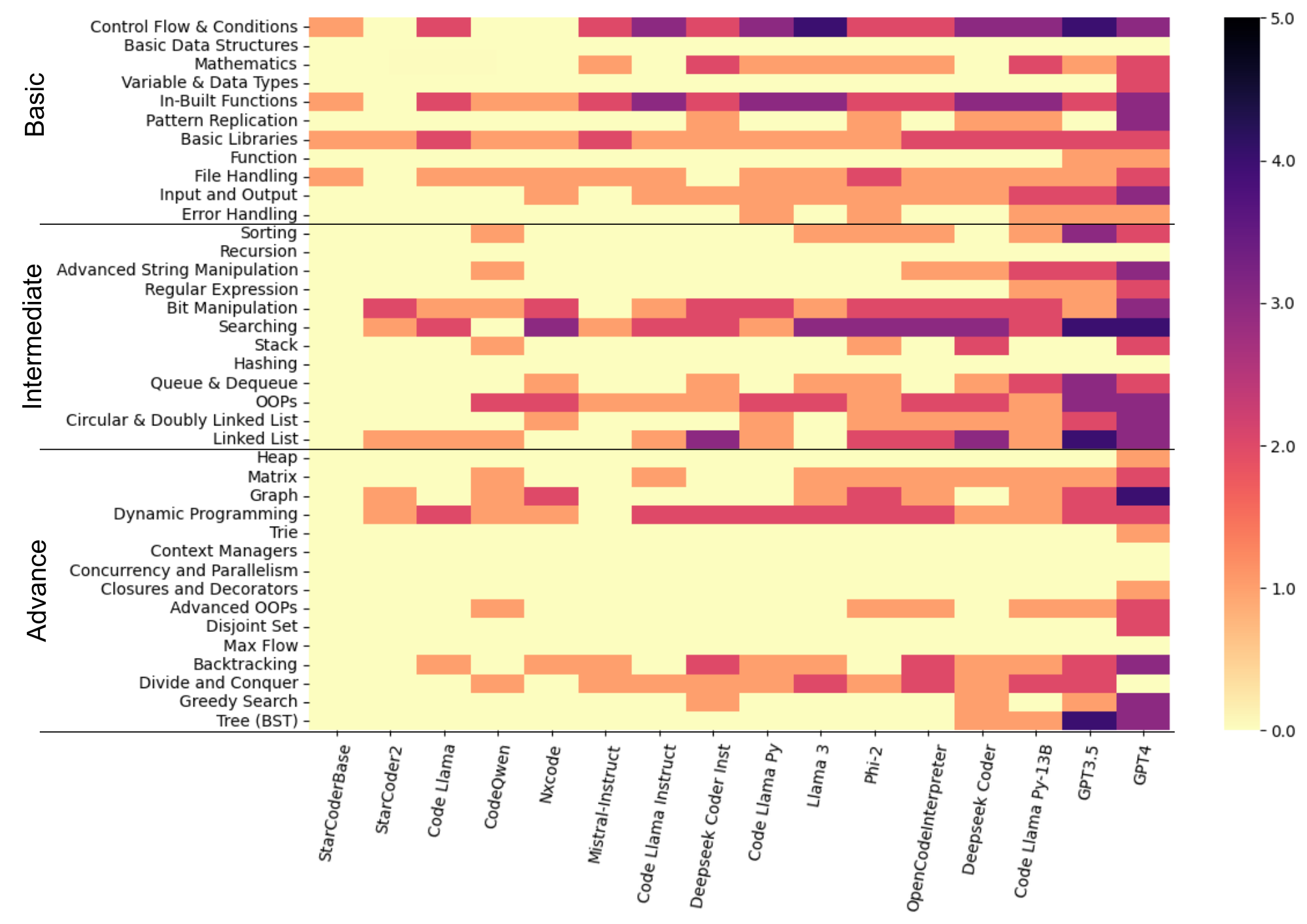}
    \caption{A heatmap showing the number of problems in \texttt{PythonSaga} solved by each LLM for a given programming concept. A model succeeds if at least one of the n(=20) generated samples passes all test cases. }\label{fig:heatmap}
\end{figure*}

Table~\ref{tab:open-close-pythonsaga} compares models against \textit{pass@1} and \textit{pass@10} metrics. Closed-source models performed considerably better than open-source models. Among open-source models, Code Llama Python~\citep{roziere2023code} 13B performed best, whereas, among closed-source models, GPT-4~\citep{openai2024gpt4} performed best. Notably, the performance of closed-source models on \texttt{PythonSaga} is significantly lower than the respective performances in HumanEval and MBPP benchmarks (see Figure~\ref{fig:temporal_performance} present in appendix for more details). 

Figure~\ref{fig:heatmap} illustrates the performance of each LLM on problems within specific programming concepts in the \texttt{PythonSaga}. We consider a model has successfully solved a problem if any one of the n(=20) generated samples passes all the test cases. As anticipated, all models exhibited better performance in solving problems associated with basic concepts compared to intermediate or advanced concepts. For example, Code Llama Python 13B solved 28.8\%, 21.6\%, and 10.9\% of problems in these categories, respectively. Whereas GPT-4 solved 42.3\%, 46.6\%, and 32.8\% of problems, respectively. In contrast to open-source models, closed-source models have successfully solved at least one problem from a majority of the concepts. Interestingly, none of the models could successfully solve any problem within six specific concepts, \textit{Basic Data Structures}, \textit{Recursion}, \textit{Hashing}, \textit{Context Managers},  \textit{Concurrancy and Parallelism} and \textit{Max Flow}. Notably, closed-source models exhibited a more consistent performance across categorization compared to open-source models, suggesting a potential advantage in handling diverse problem complexities.

\subsection{Observations}
Through our assessment of Code-LLMs using PythonSaga benchmark, we identified various error types that explain subpar performance of these LLMs. 
\begin{itemize}

    \item \textbf{Invalid syntax:} Code-LLMs frequently produce syntactically incorrect Python code, including misused keywords, unmatched parentheses, and other grammatical errors that prevent successful execution. This is one of the most common errors exhibited by all Code-LLMs during code generation.

    \item \textbf{Incomplete code:} Often, the LLMs generate incomplete code, halting mid-way through a line or statement, resulting in unusable fragments. This trait is evident in all open-source Code-LLMs but is particularly pronounced in models like StarCoder, Code Llama, Mistral, and Nxcode.

    \item \textbf{Invalid return statements:} LLMs sometimes produce return statements that are incorrect or inappropriate for the given function, leading to runtime errors or incorrect outputs. All open-source Code-LLMs occasionally produce invalid return statements, particularly when addressing advanced-level questions.

    \item \textbf{Invalid declaration of functions and variables:} LLMs sometimes incorrectly declare functions and variables, including wrong parameter lists, improper variable types, or undeclared identifiers, leading to code malfunctions. The first five Code-LLMs listed in Table~\ref{tab:open-close-pythonsaga} frequently call undeclared functions and variables or reference variables declared in prompt examples.

    \item \textbf{Subjective answers instead of code:} LLMs sometimes provide explanations or descriptions instead of generating the required code, rendering the output unusable. Models like Llama 3, Phi-2, and GPT-3.5, trained for tasks beyond code generation, are particularly prone to this error.

    \item \textbf{Copying example answers:} LLMs occasionally replicate answers provided in examples rather than generating original solutions. For complex or advanced problems, Code-LLMs like Code Llama Instruct and Deepseek Coder Instruct often use examples provided in the prompt rather than generalizing to other inputs.

    \item \textbf{Failure to pass test cases:} A significant portion of the generated code fails to pass the given test cases. This demonstrates the model's inability to produce functionally correct and robust solutions. Figure~\ref{fig:heatmap} clearly illustrates the inability of each model to pass test cases for various tasks.

    \item \textbf{Hallucination and irritable content:} Code-LLMs often generate irrelevant code and off-topic content, including random elements, facts, or links, which detract from the intended solution. All open-source Code-LLMs frequently generate URLs, often linking to GitHub, LeetCode, or other coding sites. In contrast, GPT-3 and GPT-4 exhibit this behavior almost negligibly.

    \item \textbf{Non-compliance with problem statements:} LLMs often fail to follow the format or methods outlined in problem statements, leading to invalid solutions that do not meet the specified requirements. Nearly all Code-LLMs struggle with generating compliant code when prompts become slightly complex, particularly in cases like Max Flow or Recursion.

\end{itemize}
By identifying and categorizing these errors, we highlight the need for improved benchmarks and evaluation methods that accurately reflect the capabilities and limitations of Code-LLMs.

\section{Conclusion and Future Work}
\label{sec:conclusion}
This study emphasizes the crucial need for a more balanced and comprehensive evaluation framework to ensure a fair and accurate assessment of large language models (LLMs) capable of generating code from human inputs. We address this gap by proposing an extensive categorization and hierarchy of programming concepts. An analysis of two prominent Python code generation benchmarks reveals limited diversity in both programming concepts and difficulty levels. Notably, we introduce a novel benchmark characterized by a uniform representation of concepts and difficulty, offering a more robust assessment paradigm. Our findings suggest that existing benchmarks potentially overestimate LLM performance on code generation tasks. This work lays the groundwork for the future development of diverse and representative Python code generation benchmarks, paving the way for similar studies in other programming languages.

\newpage
\pagebreak
\section*{Limitations}
This section acknowledges three key limitations associated with the present research. Firstly, due to constraints in human annotation resources, the study employed a randomly selected subset of 164 problems from the MBPP benchmark. This selection aimed to match the size of the HumanEval dataset for comparative analysis. While maintaining parity in dataset size was crucial, it is important to acknowledge that the study's findings may not generalize to the entire MBPP benchmark due to the potential for selection bias introduced by the random sampling process. Secondly, the current study employed a team of postgraduate Computer Science students with extensive experience in Python programming and competitive coding. While this selection ensured a high level of technical proficiency in the annotation task, it also acknowledges the potential limitations in terms of annotator diversity. Lastly, while the current study demonstrates the efficacy of our proposed approach within the context of the Python programming language, the generalizability of these findings to other languages requires further investigation, potentially limiting the direct applicability of our findings to benchmarks designed for languages such as Java or C++.

\section*{Ethics Statement}
 All human participants engaged in the evaluation process received detailed and comprehensible information regarding the study's nature and objectives. Prior to their involvement in the research, explicit informed consent was obtained from each participant.
 
\section*{Acknowledgements}
We express our gratitude to all the annotators for generously contributing their time and volunteering for this research without any form of remuneration. Additionally, special appreciation is extended to fellow LINGO research group\footnote{LINGO: The Computational Linguistics and Complex Social Networks Group \url{https://labs.iitgn.ac.in/lingo/}} members for their invaluable guidance and constructive feedback provided during the course of the research. 


\bibliography{anthology,custom}
\bibliographystyle{acl_natbib}

\appendix

\section{Appendix}
\label{sec:appendix}

\subsection{Performance Evaluation}
\label{appendix:performance}
Within the field of code-generating large language models (Code-LLMs), the \textit{pass@$k$} metric has emerged as a prevalent benchmark for performance evaluation~\citep{kulal2019spoc}. This metric quantifies the overall proportion of benchmark problems successfully solved by a given model. A problem is considered solved if at least one of the $k$ code samples generated by the model passes every test case associated with the problem. However, this definition leads to high variance. HumanEval~\citep{chen2021evaluating} proposed an unbiased variant, where they generate $n$ samples per problem such that n $\ge$ k, and count the number of correct samples $c \le n$ which pass unit tests. The unbiased estimator is described as:

\begin{align}
\text{pass@$k$} &:= \mathop{\mathbb{E}}_{\text{Problems}} \left[ 1 - \frac{{\binom{n-c}{k}}} {\binom{n}{k}} \right]
\label{eq:estimator}
\end{align}

Most of the Code-LLMs report \textit{pass@$k$} values at $k=1$. However, the value of $n$ varies significantly across models. For instance, STARCODER~\citep{li2023starcoder} conducts experiments with $n = 200$ for open-source models and $n=20$ for API models. 

\subsection{Representative Example from \texttt{PythonSaga}}
\label{sec:appex_rep_PythonSaga}

\begin{lstlisting}[language=Python]
{
    "task_id": "PythonSaga/15", 
    
    "prompt":
        
    def toy_distribution(n: int) -> str:
        """
        Let's say I have a bag of toys, 
        which are 'n' in number. I know 
        that these toys can be 
        distributed either to n children 
        or 1 child. 
        I want to know what can be other 
        ways to distribute these toys to 
        children in such a way that each 
        child gets at least an equal 
        number of toys.
        Take input from the user 
    `   for the number of toys. Use the 
        divmod function to solve this 
        problem.
        
        Example 1: 
        If 15 toys are there, then 15 
        children can get 1 toy each or 1 
        child can get 15 toys or 3 
        children can get 5 toys each or 
        5 children can get 3 toys each.
        In this case, 
        return 'Yes, it is possible'.
        
        Example 2: 
        If 11 toys are there, then 11 
        children can get 1 toy each or 
        1 child can get 11 toys.
        In this case, 
        return 'No, it is not possible'.
        """,
        
    "entry_point": "toy_distribution", 
        
    "canonical_solution": 
    def is_prime(n):
        """
        Check if a number is prime using 
        divmod.
        """
        if n < 2:
            return False
    
        for i in range(2,int(n**0.5)+1):
            quot,remainder=divmod(n,i)
            if remainder == 0:
                return False
    
        return True
    
    def toy_distribution(n: int) -> str:
        if n <= 0 or not is_prime(n):
            return 'Yes, it is possible'
    
        return 'No, it is not possible', 
        
    "test": 
    METADATA = {
        'author': 'AY',
        'dataset': 'test'
    }
    def check(candidate):
        assert candidate(15) == 'Yes, 
                        it is possible'
        assert candidate(11) == 'No,
                    it is not possible'
        assert candidate(20) == 'Yes,
                        it is possible'
        assert candidate(2) == 'No,
                     it is possible'
                     
            
}
\end{lstlisting}
\newpage
\subsection{Representative Example from HumanEval}
\begin{figure}[!tbh]
\begin{lstlisting}[language=Python]
{
    "task_id":"HumanEval/23", 
    
    "prompt":
        """
        def strlen(string: str) -> int:
            Return length of given 
            string   
            >>> strlen('')   
                0  
            >>> strlen('abc') 
                3""",
        
    "entry_point": "strlen", 
        
    "canonical_solution": 
        "return len(string)", 
        
    "test": 
        """METADATA = {
            'author': 'jt',
            'dataset': 'test'
            }
        def check(candidate):
           assert candidate('') == 0
           assert candidate('x') == 1
           assert candidate('asdasnakj') 
                                == 9"""
        
}
\end{lstlisting}
\caption{Representative example from the HumanEval dataset. Here, \textit{task\_id} is a unique identifier for the data sample. The \textit{prompt} contains problem text with a function header and docstrings. \textit{Canonical\_solution} presents one solution for the problem. The \textit{test} contains functions to validate the correctness of the generated code. \textit{Entry\_point} represents the function name which is yet to be completed.}

\label{fig:humaneval_example}
\end{figure}

\newpage
\subsection{Representative Example from MBPP}

\begin{figure}[!tbh]
\begin{lstlisting}[language=Python]
{
    "text": "Write a function to find m number of multiples of n.", 
    
    "code": 
        '''
        def multiples_of_num(m,n): 
           multiples_of_num= 
               list(range(n,(m+1)*n,n))
           return list(multiples_of_num)
        ''', 
    
    "task_id": 21, 
    
    "test_setup_code": "", 
    
    "test_list": 
        '''
        ["assert multiples_of_num(4,3)==
            [3,6,9,12]", 
         "assert multiples_of_num(2,5)== 
            [5,10]", 
         "assert multiples_of_num(9,2)== 
            [2,4,6,8,10,12,14,16,18]"]
         ''', 
    
    "challenge_test_list": []
}

\end{lstlisting}
\caption{Representative example from the MBPP dataset.
\textit{Text} represents the natural language description of the problem. \textit{Code} contains one possible solution for the problem. \textit{Task\_id} is the unique identifier of the sample. \textit{Test\_setup\_code} lists necessary code imports to execute tests. \textit{Test\_list} contains a list of tests to verify the solution. \textit{Challenge\_test\_list} contains a list of more challenging tests to probe the solution further.}

\label{fig:mbpp_example}
\end{figure}


\begin{figure*}[ht]
\centering
\includegraphics[trim={0.75cm 8cm 0 1cm},clip, scale=0.65]{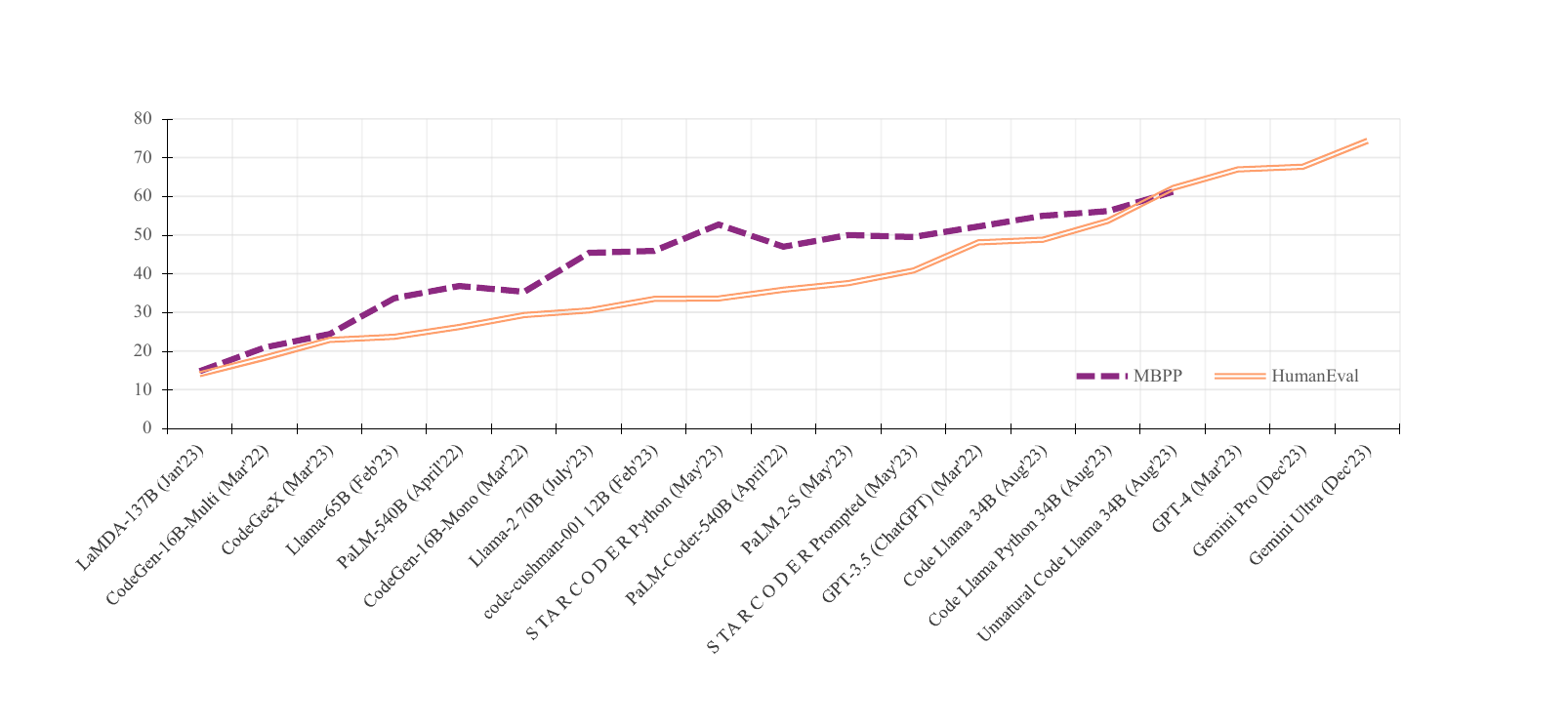}
\caption{Performance comparison arranged in ascending order of (pass@1) of popular Code-LLMs on two Python benchmarks, HumanEval~\citep{chen2021evaluating} and MBPP~\citep{austin2021program}. \textit{pass@1} scores are taken verbatim as reported in STARCODER~\citep{li2023starcoder}, Code Llama~\citep{roziere2023code}, and Gemini~\citep{team2023gemini}. GPT-4, Gemini Pro, and Gemini Ultra do not report performance scores on MBPP dataset.}\label{fig:temporal_performance}
\end{figure*}


\end{document}